\documentclass[letterpaper]{article} % DO NOT CHANGE THIS
\usepackage{aaai2026}  % DO NOT CHANGE THIS
\usepackage{times}  % DO NOT CHANGE THIS
\usepackage{helvet}  % DO NOT CHANGE THIS
\usepackage{courier}  % DO NOT CHANGE THIS
\usepackage[hyphens]{url}  % DO NOT CHANGE THIS
\usepackage{graphicx} % DO NOT CHANGE THIS
\usepackage{natbib}  % DO NOT CHANGE THIS AND DO NOT ADD ANY OPTIONS TO IT
\usepackage{caption} % DO NOT CHANGE THIS AND DO NOT ADD ANY OPTIONS TO IT
\usepackage{algorithm}
\usepackage{algorithmic}

\usepackage{newfloat}
\usepackage{listings}
\DeclareCaptionStyle{ruled}{labelfont=normalfont,labelsep=colon,strut=off} % DO NOT CHANGE THIS
\floatstyle{ruled}
\newfloat{listing}{tb}{lst}{}
\floatname{listing}{Listing}
\title{Pre-DPO: Improving Data Utilization in Direct Preference Optimization\\Using a Guiding Reference Model}
\author{
    Junshu Pan\textsuperscript{\rm 1,\rm 2,\rm 3},
    Wei Shen\textsuperscript{\rm 4},
    Shulin Huang\textsuperscript{\rm 1,\rm 2},
    Qiji Zhou\textsuperscript{\rm 2},
    Yue Zhang\textsuperscript{\rm 2}\thanks{Corresponding author.}
}
\affiliations{
    \textsuperscript{\rm 1}Zhejiang University\\
    \textsuperscript{\rm 2}School of Engineering, Westlake University\\
    \textsuperscript{\rm 3}Shanghai Innovation Institute\\
    \textsuperscript{\rm 4}Independent Researcher\\
    \{panjunshu,zhangyue\}@westlake.edu.cn
}

\usepackage{bibentry}
\usepackage{booktabs}
\usepackage{multirow}
\usepackage{amsmath}
\usepackage{amssymb}

\begin{document}

\maketitle

\begin{abstract}
Direct Preference Optimization (DPO) simplifies reinforcement learning from human feedback (RLHF) for large language models (LLMs) by directly training on offline preference data to align with human preferences.
During DPO training, the reference model serves as a data weight adjuster.
However, the common practice of initializing the policy and reference models identically in DPO can lead to inefficient data utilization and impose a performance ceiling.
Meanwhile, the absence of a reference model in Simple Preference Optimization (SimPO) reduces training robustness and requires stricter conditions to prevent catastrophic forgetting.
In this work, we propose \textbf{Pre-DPO}, a simple yet effective DPO-based training paradigm that improves preference optimization by introducing a \textit{guiding reference model}.
This reference model provides \textit{foresight} into the desired policy state achievable through the training preference data, serving as a guiding mechanism that adaptively assigns higher weights to samples more suitable for the model and lower weights to those less suitable.
Extensive experiments on the AlpacaEval 2 and Arena-Hard v0.1 benchmarks demonstrate that Pre-DPO consistently improves the performance of both DPO and SimPO, without relying on external models or additional data.
\end{abstract}

\begin{links}
    \link{Code}{https://github.com/DtYXs/Pre-DPO}
    % \link{Extended version}{https://arxiv.org/abs/2504.15843}
\end{links}

\section{Introduction}

Preference-based training has become a widely adopted and effective paradigm for aligning large language models (LLMs) with human values and preferences. Direct Preference Optimization (DPO)~\cite{rafailov2023direct}, as a representative of reference-based preference optimization methods~\cite{rafailov2023direct,ethayarajh2024kto,azar2024general,gorbatovski2025learn}, directly trains LLMs on preference data under the constraint of a reference model, without relying on an explicit reward model or complex online reinforcement learning.

\begin{figure}[t]
    \centering
    \includegraphics[width=0.95\columnwidth]{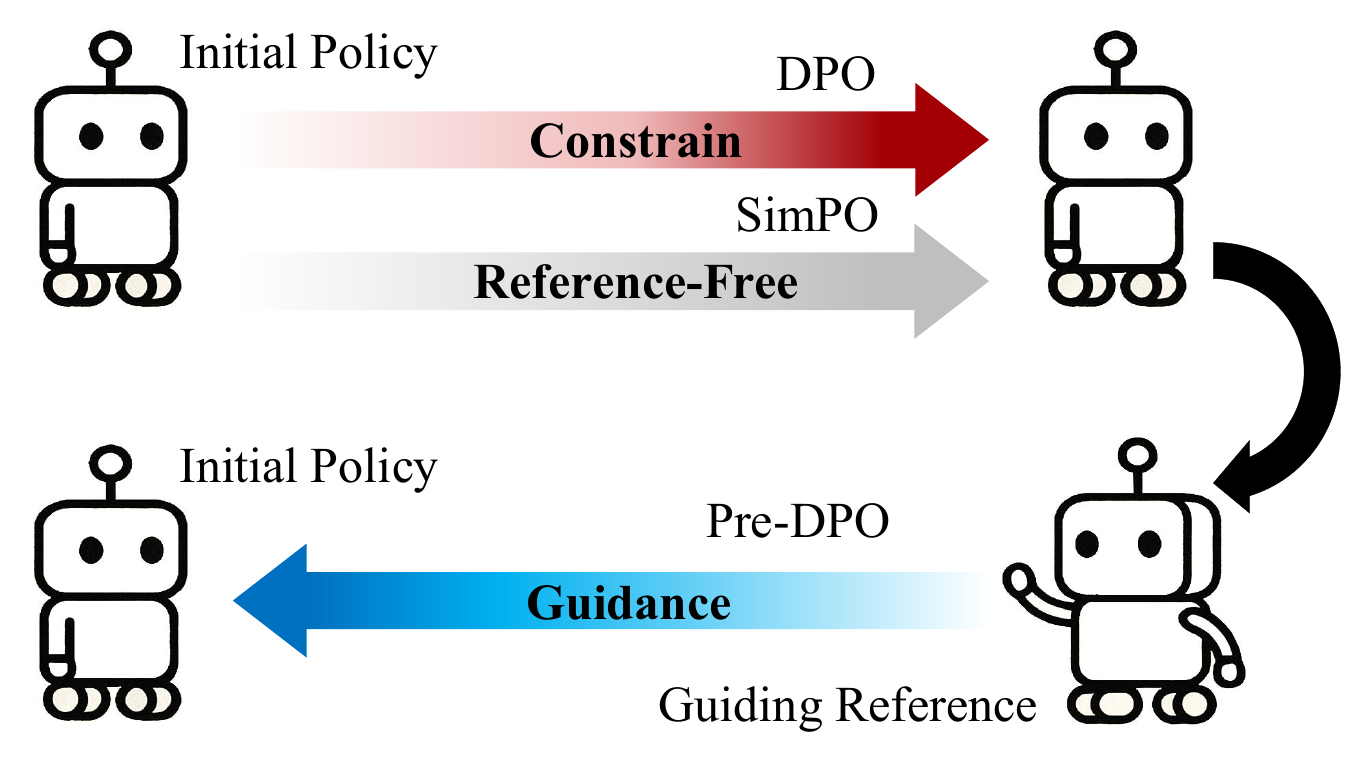}
    \caption{Pre-DPO introduces a guiding reference model derived from the optimized policy to guide re-optimization, transforming the reference from a constraint into an informed guide with foresight.}
    \label{fig:figure1}
\end{figure}

Recently, it has been shown that a reference model is not necessary for achieving effective preference optimization. Simple Preference Optimization (SimPO)~\cite{meng2024simpo}, as a representative of reference-free preference optimization methods (\citealp{xu2023some}; \citealp{xu2024contrastive}; \citealp{hong2024orpo}; \citealp{meng2024simpo}; \citealp{nath2025dpl}), eliminates the need for a reference model and yields better performance and efficiency, though at the cost of an increased risk of catastrophic forgetting~\cite{meng2024simpo}. Meanwhile, other studies have empirically demonstrated that DPO can benefit either from relaxing the constraints imposed by the reference model~\cite{gorbatovski2025learn} or from stronger external reference models~\cite{liu2024understanding}. However, decreasing reliance on a reference model imposes stricter practical requirements to ensure effective learning, and clear methodologies for obtaining an appropriate reference model are still lacking.

Despite the empirical efforts, the role of the reference model in DPO and its impact on training dynamics remain insufficiently explored. During DPO training, the reference model serves as a data weight adjuster (see Section~\ref{motivation}). It adaptively tends to assign higher weights to data aligned with itself while reducing weights for conflicting data. However, due to the common practice in DPO of initializing the policy and reference models identically~\cite{rafailov2023direct}, as training progresses, the reference model increasingly constrains the policy model by penalizing deviations, potentially introducing a performance ceiling. Moreover, identical initialization of policy and reference models results in the nearly uniform weighting of training examples during the early training stages, which is in contrast to prior studies showing that assigning non-uniform weights to training data can lead to improved learning and performance~\cite{lin2017focal,ren2018learning,shu2019meta}.

In light of the limitations of conventional reference models, we hypothesize that an ideal reference model for DPO should originate from the initial policy model and provide insights into potential directions for policy improvement based on the preference data. We define this type of reference model as a \textit{guiding reference model}, which can better support learning by transforming the role of the reference model from a constraint into a guide with \textit{foresight} (see Section~\ref{foresight}). Building on this insight, we propose Pre-DPO, a simple yet effective training paradigm that enhances data utilization and improves the performance of existing preference optimization methods without relying on external models or additional data, as shown in Figure~\ref{fig:figure1}. Pre-DPO first optimizes the initial policy using a standard preference optimization method (e.g., DPO or SimPO). The resulting optimized policy is then employed as the guiding reference model. Finally, the initial policy is re-optimized using DPO with this guiding reference model, yielding a better-optimized policy through more effective data reweighting. The guiding reference model in Pre-DPO essentially serves as an adaptive guiding mechanism that naturally assigns higher weights to samples more suitable for the model and lower weights to those less suitable. In practice, these suitable cases typically correspond to examples that are easier to learn, allowing the model to efficiently leverage data that aligns well with its learning trajectory.

We evaluate Pre-DPO on the Llama3.2-3B~\cite{grattafiori2024llama} and Qwen2.5-7B~\cite{yang2024qwen2} model series across AlpacaEval 2~\cite{alpaca_eval, dubois2024length} and Arena-Hard v0.1~\cite{li2024crowdsourced} benchmarks. The experimental results show that Pre-DPO consistently improves the performance of both DPO and SimPO, achieving average gains of 2.5 points in length-controlled win rate (LC) on AlpacaEval 2 and 2.6 points in win rate (WR) on Arena-Hard v0.1. By introducing the guiding reference model, Pre-DPO can further improve the performance of existing well-tuned preference optimization methods, effectively overcoming the performance ceiling caused by inefficient data utilization under traditional reference model settings. Notably, Pre-DPO does not rely on external models or additional data, making it highly flexible and easy to deploy.

\section{Related Work}

Reinforcement learning from human feedback (RLHF)~\cite{christiano2017deep,stiennon2020learning} has become an effective approach for aligning LLMs with human values and preferences~\cite{ouyang2022training,achiam2023gpt, grattafiori2024llama,yang2024qwen2,shen2025exploring}. Typically, pretrained LLMs first undergo supervised fine-tuning (SFT) to learn instruction-following behavior. Subsequently, reinforcement learning is conducted using external reward models and policy optimization algorithms~\cite{schulman2017proximal,shao2024deepseekmath,hu2025reinforce++}, which are typically performed online. While RLHF has demonstrated strong performance, its full pipeline remains complex and resource-intensive. DPO~\cite{rafailov2023direct} simplifies the RLHF process by directly optimizing on preference data in an offline setting. In this work, we focus on improving data utilization in such offline preference optimization.

Offline preference optimization eliminates the need for an explicit reward model and avoids the complex online learning optimization process. A well-optimized model obtained via offline preference optimization can also serve as a strong initial policy for subsequent online reinforcement learning optimization~\cite{yang2024qwen2}. Depending on whether or not a reference model is required, offline preference optimization can be classified into two categories: reference-based preference optimization methods~\cite{rafailov2023direct,ethayarajh2024kto,azar2024general,kim2025sdpo,gorbatovski2025learn} and reference-free preference optimization methods (\citealp{xu2023some}; \citealp{xu2024contrastive}; \citealp{hong2024orpo}; \citealp{meng2024simpo}; \citealp{nath2025dpl}). DPO~\cite{rafailov2023direct}, as a representative of reference-based preference optimization methods, directly trains on preference data and implicitly optimizes the same objective as existing reinforcement learning algorithms. SimPO~\cite{meng2024simpo}, as a representative of reference-free preference optimization methods, removes the need for a reference model and can achieve better results than DPO at the cost of lower training robustness~\cite{meng2024simpo}. In this paper, we experimentally demonstrate that DPO can also benefit from reference-free preference optimization methods by leveraging a guiding reference model.

For the reference models in DPO, prior work~\cite{liu2024understanding} empirically suggested that DPO can benefit from a stronger and more suitable reference model in certain cases. However, it mainly focuses on stronger external reference models and does not provide a theoretical explanation for why they can be beneficial. \citet{gorbatovski2025learn} proposed a dynamic update strategy to reset the reference model based on the current policy, which could weaken the regularization effect of the reference model and tends to assign more identical weights to the data samples with more frequent updates. In this work, we introduce the concept of a guiding reference model, analyze its role in enhancing data utilization in DPO through better data reweighting, and propose an effective methodology for leveraging it in the DPO framework.

\section{Preliminaries}

Given a text prompt $x$, the RLHF training stage aims to increase the probability that an LLM generates a response $y$ that is better aligned with human values. Specifically, the objective is to maximize the expected reward $r(x,y)$ while controlling the deviation between the policy probability distribution $\pi_\theta(y \mid x)$ and the reference probability distribution $\pi_\text{ref}(y \mid x)$. The optimization objective is formulated as follows:
\begin{equation}
    \max_{\pi_{\theta}} \mathbb{E}_{(x, y) \sim \mathcal{D} \times \pi_\theta} \left[ r(x, y) \right] - \beta \mathbb{D}_{\text{KL}} \left[ \pi_\theta(y|x) \parallel \pi_{\text{ref}}(y|x) \right],
    \label{eq:RL}
\end{equation}
where $\mathcal{D}\times \pi_\theta$ denotes the joint distribution of the prompt $x$ and the response $y$ from $\pi_\theta(y \mid x)$. The KL divergence term constrains the deviation of the policy $\pi_\theta$ from the reference model, and $\beta$ controls the strength of this constraint.

DPO~\cite{rafailov2023direct} is a widely used reference-based preference optimization method that eliminates the need for explicit reward signals and has become a component in the post-training pipeline of many popular open-source LLMs~\cite{bi2024deepseek,jiang2024mixtral,yang2024qwen2,xu2025qwen25omni}. It reformulates Eq. \ref{eq:RL} into a direct optimization process on a preference dataset $\mathcal{D} = \{(x_i, y^+_i, y^-_i)\}_{i=1}^{|\mathcal{D}|}$, where $x$ is the prompt, $y_i^+$ is the preferred response, and $y_i^-$ is the less-preferred response. The objective of DPO is as follows:
\begin{align}
    \mathcal{L}_{\text{DPO}}(\pi_{\theta}; \pi_{\text{ref}}) \! = \!
    & -\mathbb{E}_{(x, y^+, y^-) \sim \mathcal{D}} \Big[ \log \sigma \Big( 
    \beta \log \frac{\pi_{\theta}(y^+ \mid x)}{\pi_{\text{ref}}(y^+ \mid x)} \nonumber \\
    & \qquad \qquad - \beta \log \frac{\pi_{\theta}(y^- \mid x)}{\pi_{\text{ref}}(y^- \mid x)} 
    \Big) \Big]
    \label{eq:DPO loss}
\end{align}
where $\sigma(\cdot)$ is the sigmoid function, and $\beta$ is a hyperparameter that controls the strength of the reference model's constraint.

A large $\beta$ imposes a strong constraint, which can limit the model's performance improvement, while a small $\beta$ may result in insufficient constraints, leading to model degradation~\cite{liu2024understanding}. In practical LLM training, $\pi_\theta$ and $\pi_\text{ref}$ are typically initialized as the supervised fine-tuning (SFT) model. $\pi_\text{ref}$ remains fixed during training.

SimPO~\cite{meng2024simpo} is a reference-free preference optimization method that removes the reference model in DPO while introducing length normalization and target reward margin. Its loss function is defined as:

\begin{align}
\mathcal{L}_{\text{SimPO}}(\pi_{\theta}) = 
& -\mathbb{E}_{(x, y^+, y^-) \sim \mathcal{D}} \Big[ \log \sigma \Big( \nonumber \\
& \!\!\!\!\!\!\!\!\!\!\!\!\!\!\!\!\! \frac{\beta}{|y^+|} \log \pi_{\theta}(y^+ \mid x) 
- \frac{\beta}{|y^-|} \log \pi_{\theta}(y^- \mid x) - \gamma 
\Big) \Big]
\label{eq:SimPO loss}
\end{align}
where $|y^+|$ and $|y^-|$ denote the lengths, $\beta$ is a hyperparameter constant, and $\gamma$ is the target reward margin. SimPO has the potential to surpass DPO in performance but suffers from reduced robustness due to the lack of reference constraints~\cite{meng2024simpo}.

\section{Method}

\begin{figure*}[t]
    \centering
    \includegraphics[width=0.95\textwidth]{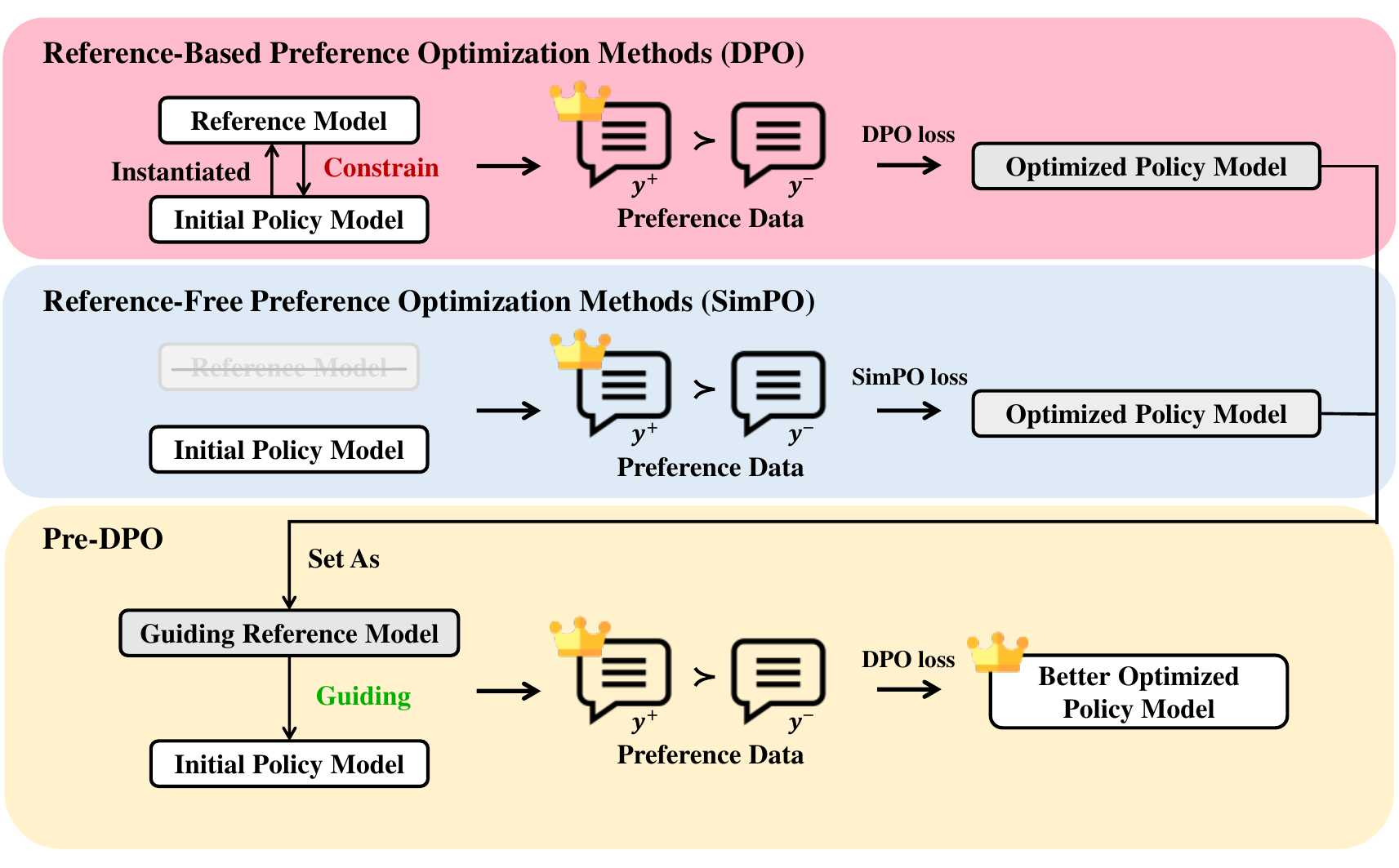}
    \caption{An overview of Pre-DPO. DPO constrains training using the initial policy model as the reference, while SimPO is reference-free. Pre-DPO first optimizes a policy model using DPO or SimPO, then resets it as a guiding reference model, and re-optimizes the initial policy using DPO. This process enhances data utilization and results in a better-optimized policy model.}
    \label{fig:Pre-DPO}
\end{figure*}

In this section, we first present the motivation behind Pre-DPO, focusing on the limitations of the reference setting in DPO and the specific challenges Pre-DPO addresses. Then, we provide a detailed explanation of Pre-DPO, outlining how it adaptively reweights the training examples and describing its overall process.

% \subsection{Limitations of the Reference Model for Data Reweighting in Vanilla DPO}
\subsection{Challenges in Data Reweighting with Vanilla DPO’s Reference Model}

\label{motivation}
From the loss function of DPO (Eq. \ref{eq:DPO loss}), we can derive the gradient with respect to the parameters $\theta$ (the detailed derivation is provided in Appendix~\ref{sec:appendix_derivation}):
\begin{align}
    \nabla_\theta \mathcal{L}_{\text{DPO}}(\pi_\theta; \pi_{\text{ref}}) = \nonumber \\ 
    & \!\!\!\!\!\!\!\!\!\!\!\!\!\!\!\!\!\!\!\!\!\!\!\!\!\!\!\!\!\! - \beta \mathbb{E}_{(x, y^+, y^-) \sim \mathcal{D}} \left[ 
    \lambda \cdot
    \nabla_\theta \log \frac{\pi_\theta(y^+ \mid x)}{\pi_\theta(y^- \mid x)}
    \right],
\end{align}
where $\lambda$ is defined as:
\begin{equation}
    \lambda = \sigma\left( \beta \log \frac{\pi_{\text{ref}}(y^+|x)}{\pi_{\text{ref}}(y^-|x)} - \beta \log \frac{\pi_{\theta}(y^+|x)}{\pi_{\theta}(y^-|x)} \right).
    \label{eq:lambda}
\end{equation}
From the perspective of example reweighting, DPO learns from preference pairs with weights $\lambda$, where the reference model $\pi_\text{ref}$ controls the training process by adjusting $\lambda$.

When the policy $\pi_\theta$ and the reference $\pi_{\text{ref}}$ are initialized from an identical SFT model, $\lambda$ starts around the constant 0.5 in the early stage of training due to $\sigma(0) = 0.5$. However, a constant $\lambda$ can lead to degeneracy~\cite{rafailov2023direct}, and more importantly, previous research in the field of example reweighting has shown that assigning appropriate and varying weights to training samples can improve model performances and data efficiency~\cite{lin2017focal,ren2018learning,shu2019meta}.

On the other hand, as training progresses, the reference continuously constrains the policy deviation by adjusting the value of $\lambda$. Specifically, when $\frac{\pi_{\text{ref}}(y^+|x)}{\pi_{\text{ref}}(y^-|x)}$ is large, it encourages a larger value of $\lambda$, promoting learning from the corresponding preference pair. Conversely, a smaller ratio typically leads to a lower $\lambda$, which in turn reduces the influence of that sample on the learning process. The difference in $\lambda$ between DPO and Pre-DPO is largely driven by the reference model, particularly in the early stage of training. Therefore, a suboptimally configured reference model can result in suboptimal weighting of training samples.

\subsection{Pre-DPO: Improving Data Utilization in DPO Using a Guiding Reference Model}
\label{foresight}

One straightforward solution is to employ a reference model that differs from the initial policy and provides foresight into promising directions for policy improvement based on the preference data $\mathcal{D}$, enabling more effective data reweighting and guidance during training.

Notably, a model that has already undergone preference optimization contains information about the entire training process. More importantly, it reflects the outcomes that the initial policy can achieve through the available preference data. Specifically, when the reference model in DPO is set to a guiding reference model $\pi_{\text{guide}}$, the weight $\lambda$ becomes:
\begin{equation}
    \lambda = \sigma\left( \beta \log \frac{\pi_{\text{guide}}(y^+|x)}{\pi_{\text{guide}}(y^-|x)} - \beta \log \frac{\pi_{\theta}(y^+|x)}{\pi_{\theta}(y^-|x)} \right).
    \label{eq:lambda_guide}
\end{equation}

The foresight of the guiding reference model is reflected in its way to modulate $\lambda$: assigning higher weights to samples that the policy model can learn effectively, while down-weighting those that are difficult to learn or potentially conflicting. This behavior naturally aligns with findings suggesting that avoiding ambiguous, mislabeled, or overly difficult preference data can benefit alignment~\cite{houliston2024uncertainty, gao2025principled}. Compared to the reference model used in standard DPO, which merely serves as a constraint without foresight, the guiding reference model enables more informed and data-dependent reweighting, leading to more efficient and targeted policy improvement.

Hence, we propose a simple yet effective paradigm for obtaining a suitable reference without needing external models. Specifically, we reset the optimized policy model as the reference for the next training iteration. Since the reference model retains all the information from prior policy training, its role shifts from a constraint to that of a guide, which we refer to as the \textit{guiding reference model}. Employing this guiding reference model in DPO adaptively assigns higher weights to training data that aligns with it while reducing the weights of conflicting samples.

Let $\pi_\theta$ denote the policy model to be optimized, $\pi_\text{ref}$ represent the reference model (which can also be set as None), and $\mathcal{M}(\pi_\theta;\pi_\text{ref})$ indicate the preference optimization method. The procedure of Pre-DPO (illustrated in Figure~\ref{fig:Pre-DPO}) is described in detail below:

\paragraph{Step 1: Instantiate the initial reference model.}  
If $\mathcal{M}$ is reference-based, set the reference $\pi_{\text{ref}}$ to $\pi_{\text{SFT}}$. Otherwise, for reference-free methods, $\pi_{\text{ref}}$ is set to \texttt{None}:
\begin{equation}
    \pi_{\text{ref}} = \begin{cases} 
    \pi_{\text{SFT}} & \text{for reference-based $\mathcal{M}$}, \\
    \texttt{None} & \text{for reference-free $\mathcal{M}$}.
    \end{cases}
\end{equation}

\paragraph{Step 2: The first round of preference optimization.}  
Perform preference optimization $\mathcal{M}$ on $\pi_\text{SFT}$ with the preference dataset $\mathcal{D}$:
\begin{equation}
    \pi_{\mathcal{M}} = \mathcal{M}(\pi_\text{SFT}; \pi_\text{ref}).
\end{equation}

\paragraph{Step 3: Set the guiding reference model.}
After the first round of optimization, reset $\pi_{\text{ref}}$ to the optimized model $\pi_{\mathcal{M}}$ obtained from the previous round. This optimized model now serves as the guiding reference model $\pi_{\text{guide}}$:
\begin{equation}
    \pi_\text{guide} = \pi_{\mathcal{M}}.
\end{equation}

\paragraph{Step 4: Preference optimization with the guiding reference model.}  
Apply DPO to $\pi_{\text{SFT}}$ using the guiding reference $\pi_{\mathcal{\text{guide}}}$ on the same dataset $\mathcal{D}$ to obtain the better optimized model $\pi_{\text{Pre-DPO}}$:
\begin{equation}
    \pi_{\text{Pre-DPO}} = \mathcal{M}_{\text{DPO}}(\pi_{\text{SFT}}; \pi_\text{guide}).
\end{equation}

\section{Experiments}

We empirically evaluate the effectiveness of Pre-DPO in enhancing existing preference optimization methods through a guiding reference model. To ensure a comprehensive and fair assessment, we conduct experiments on the Llama3.2-3B~\cite{grattafiori2024llama} and Qwen2.5-7B~\cite{yang2024qwen2} model series, including both Base and Instruct versions. We evaluate the models' performance on two widely-used preference optimization benchmarks: AlpacaEval 2~\cite{alpaca_eval,dubois2024length} and Arena-Hard v1.0~\cite{li2024crowdsourced}. Given the sensitivity of preference optimization to hyperparameters~\cite{meng2024simpo}, we conduct an extensive hyperparameter search to ensure reliable results. All experiments are conducted based on the LlamaFactory~\cite{zheng2024llamafactory} repository, and all models and datasets used are publicly available.

\subsection{Experimental Setup}
\paragraph{Models and datasets.}
In our experiments, we primarily consider two widely recognized series of open-source models, Llama3.2-3B and Qwen2.5-7B, including both Base and Instruct versions. The diversity in model types and scales enables a more comprehensive evaluation of our method’s effectiveness.

For Base models, we first train them on the UltraChat-200k~\cite{ding2023enhancing} dataset to obtain their corresponding SFT models. We then directly perform preference optimization on the existing binarized UltraFeedback~\cite{cui2024ultrafeedbackboostinglanguagemodels} dataset, which is widely used in prior work on offline preference optimization training~\cite{ethayarajh2024kto,hong2024orpo,meng2024simpo,liu2024understanding,kim2025sdpo,gorbatovski2025learn}, using the SFT model as the initialization.

For Instruct models, we directly use them as SFT models. During the preference optimization stage, we construct two additional on-policy preference datasets for Llama3.2-3B-Instruct and Qwen2.5-7B-Instruct, respectively. Specifically, for each prompt in the binarized UltraFeedback dataset, we sample six responses from each Instruct model using a temperature of 0.8 and a top-$p$ of 0.95 as sampling parameters. Subsequently, following prior work\cite{meng2024simpo}, we use the ArmoRM-Llama3-8B-v0.1~\cite{wang2024interpretable} reward model to score each response and select the highest-scoring and lowest-scoring responses to form preference pairs $(x, y^+, y^-)$. We discard prompts for which all sampled responses receive identical scores. This process results in two new preference datasets, one for each of the Llama3.2-3B-Instruct and Qwen2.5-7B-Instruct models.

\paragraph{Baselines.}
Prior work~\cite{meng2024simpo} shows that, with sufficient hyperparameter tuning, both DPO~\cite{rafailov2023direct} and SimPO~\cite{meng2024simpo} are highly competitive. Therefore, we adopt them as representative baselines for the reference-based and reference-free methods in our large-scale main experiments. Additionally, we conduct experiments with ORPO~\cite{hong2024orpo}, sDPO~\cite{kim2025sdpo} and TR-DPO~\cite{gorbatovski2025learn} under the Llama3.2-3B-Base setting.
% , and with Iterative DPO~\cite{yuan2024self} under the Llama3.2-3B-Instruct setting. 

\paragraph{Implementation details.}
\label{sec:implementation_details}
For the SFT stage in Base models, we train for 3 epochs using a batch size of 32, a maximum sequence length of 4096, a learning rate of $2 \times 10^{-6}$, and a cosine learning rate schedule with a 6\% warmup ratio.

All preference optimization experiments use a batch size of 128, a maximum sequence length of 4096, and a cosine learning rate schedule with a 6\% warmup ratio, training for 1 epoch. All models are fine-tuned using full parameter tuning.

Given the importance of hyperparameter tuning in offline preference optimization~\cite{meng2024simpo}, we perform extensive hyperparameter searches for all preference optimization experiments to ensure fairness. Specifically, for the key hyperparameters, including the learning rate, $\beta$ (for DPO-based methods and SimPO), and $\gamma$ (for SimPO), a two-stage tuning strategy is employed. We first fix the learning rate and search for the optimal $\beta$ or $\gamma$. Then, with the best $\beta$ or $\gamma$ fixed, we search for the optimal learning rate. More details of hyperparameter tuning and the best hyperparameter setting can be found in Appendix~\ref{sec:appendix_implementation_details}.

\paragraph{Evaluation benchmarks.}
We evaluate methods primarily on two open-source instruction-following benchmarks: AlpacaEval 2~\cite{alpaca_eval, dubois2024length} and Arena-Hard v0.1~\cite{li2024crowdsourced}, which are widely adopted in the community for evaluating the instruction-following capabilities of LLMs~\cite{meng2024simpo}. We report the raw win rate (WR) and length-controlled win rate (LC) on AlpacaEval 2, and the WR on Arena-Hard v0.1, using their respective official repositories. More evaluation details can be found in Appendix~\ref{sec:appendix_evaluation_details}.

\subsection{Main Results}

\begin{table*}[t]
\centering
\setlength{\tabcolsep}{1.36mm}
\begin{tabular}{llcccccccc}
\toprule
\multirow{4}{*}{\textbf{Method}} & \multirow{4}{*}{\textbf{Ref.}} & \multicolumn{4}{c}{\textbf{Llama3.2-3B-Base}} & \multicolumn{4}{c}{\textbf{Llama3.2-3B-Instruct}} \\ 
\cmidrule(lr){3-6}\cmidrule(lr){7-10} & & \multicolumn{3}{c}{\textbf{AlpacaEval 2}} & \multicolumn{1}{c}{\textbf{Arena-Hard}} & \multicolumn{3}{c}{\textbf{AlpacaEval 2}} & \multicolumn{1}{c}{\textbf{Arena-Hard}} \\ 
\cmidrule(lr){3-5} \cmidrule(lr){6-6} \cmidrule(lr){7-9}\cmidrule(lr){10-10} & & {\small \bf LC (\%)} & {\small \bf WR (\%)} & \small \bf Len. & {\small \bf WR (\%)} & {\small \bf LC (\%)} & {\small \bf WR (\%)} & \small \bf Len. & {\small \bf WR (\%)} \\
\midrule
SFT & - & 6.1 & 4.0 & 1012 & 2.1 & 19.0 & 18.9 & 1956 & 18.5 \\
\midrule
DPO & SFT & 10.5 & 12.0 & 2042 & 10.6 & 36.3 & 36.9 & 2026 & 30.6 \\
Pre-DPO & DPO & 12.5 \small{(+19.0\%)} & 13.9 \small{(+15.8\%)} & 2061 & 11.9 \small{(+12.3\%)} & \textbf{39.3} \small{(+8.3\%)} & \textbf{40.9} \small{(+10.8\%)} & 2095 & \textbf{34.7} \small{(+13.4\%)} \\
\midrule
SimPO & - & 13.1 & 13.1 & 1950 & 11.7 & 33.8 & 29.9 & 1797 & 28.1 \\
Pre-DPO & SimPO & \textbf{18.1} \small{(+38.2\%)} & \textbf{18.4} \small{(+40.5\%)} & 2020 & \textbf{14.0} \small{(+19.7\%)} & 35.0 \small{(+3.6\%)} & 32.3 \small{(+8.0\%)} & 1846 & 30.0 \small{(+6.8\%)} \\
\midrule[.7pt]
\multirow{4}{*}{\textbf{Method}} & \multirow{4}{*}{\textbf{Ref.}} & \multicolumn{4}{c}{\textbf{Qwen2.5-7B-Base}} & \multicolumn{4}{c}{\textbf{Qwen2.5-7B-Instruct}} \\ 
\cmidrule(lr){3-6}\cmidrule(lr){7-10} & & \multicolumn{3}{c}{\textbf{AlpacaEval 2}} & \multicolumn{1}{c}{\textbf{Arena-Hard}} & \multicolumn{3}{c}{\textbf{AlpacaEval 2}} & \multicolumn{1}{c}{\textbf{Arena-Hard}} \\ 
\cmidrule(lr){3-5}\cmidrule(lr){6-6} \cmidrule(lr){7-9} \cmidrule(lr){10-10} & & {\small \bf LC (\%)} & {\small \bf WR (\%)} & \small \bf Len. & {\small \bf WR (\%)} & {\small \bf LC (\%)}  & {\small \bf WR (\%)} & \small \bf Len. & {\small \bf WR (\%)} \\
\midrule
SFT & - & 18.6 & 6.9 & 892 & 9.4 & 31.2 & 31.0 & 2020 & 55.9 \\
\midrule
DPO & SFT & 24.8 & 22.2 & 1778 & 33.1 & 52.2 & 56.8 & 2270 & 62.9 \\
Pre-DPO & DPO & 27.4 \small{(+10.5\%)} & 24.5 \small{(+10.4\%)} & 1790 & 32.6 \small{(-1.5\%)} & 53.3 \small{(+2.1\%)} & \textbf{59.4} \small{(+4.6\%)} & 2322 & \textbf{68.8} \small{(+9.4\%)} \\
\midrule
SimPO & - & 34.7 & 31.9 & 1836 & 38.1 & 51.7 & 52.4 & 2119 & 62.4 \\
Pre-DPO & SimPO & \textbf{37.2} \small{(+7.2\%)} & \textbf{32.6} \small{(+2.2\%)} & 1758 & \textbf{41.6} \small{(+9.2\%)} & \textbf{54.6} \small{(+5.6\%)} & 55.5 \small{(+5.9\%)} & 2121 & 64.5 \small{(+3.4\%)} \\
\bottomrule
\end{tabular}
\caption{Performance of Pre-DPO under four different model settings on AlpacaEval 2 and Arena-Hard v0.1. LC and WR denote the length-controlled and raw win rate, respectively. Ref. denotes the reference model and Len. denotes the average response length. The SFT models for the Base settings are trained on the UltraChat-200k dataset, while the Instruct models are used as the SFT models directly for the Instruct settings. The guiding reference models are obtained from DPO and SimPO.}
\label{table:main_result}
\end{table*}

% table: ablation
\begin{table}[t]
\centering
\small
\setlength{\tabcolsep}{0.25mm}
\begin{tabular}{lllcccc}
\toprule
\multirow{3}{*}{\textbf{}} & \multirow{3}{*}{\textbf{Method}} & \multirow{3}{*}{\textbf{Ref.}} & \multirow{3}{*}{\textbf{Epoch}} & \multicolumn{3}{c}{\textbf{AlpacaEval 2}} \\ 
\cmidrule(lr){5-7} 
& & & & {\small \textbf{LC (\%)}} & {\small \textbf{WR (\%)}} & {\small \textbf{Len.}} \\
\midrule
\multirow{3}{*}{Base}     & DPO       & SFT              & 1 & 10.5          & 12.0          & 2042 \\
\cmidrule(lr){2-7}
                                      & DPO       & SFT              & 2 & 11.0 \small{(+4.8\%)}          & 12.0 \small{(+0.0\%)}         & 1976 \\
                                      & Pre-DPO & $\text{DPO}_{1}$ & 1 & \textbf{12.5} \small{(+19.0\%)} & \textbf{13.9} \small{(+15.8\%)} & 2061 \\
\midrule
\multirow{3}{*}{Instruct} & DPO       & SFT              & 1 & 36.3          & 36.9          & 2026 \\
\cmidrule(lr){2-7}
                                      & DPO       & SFT              & 2 & 35.2 \small{(-3.0\%)}          & 37.1 \small{(+0.5\%)}         & 2113 \\
                                      & Pre-DPO & $\text{DPO}_{1}$ & 1 & \textbf{39.3} \small{(+8.3\%)} & \textbf{40.9} \small{(+10.8\%)} & 2095 \\
\bottomrule 
\end{tabular}
\caption{Ablation studies under the Llama3.2-3B model settings. DPO trained for 2 epochs has the same computational cost as Pre-DPO. $\text{DPO}_1$ denote the guiding reference model trained with DPO for 1 epoch.}
\label{table:ablation_ref}
\end{table}

% table: other_result_more
\begin{table}[t]
\centering
\small
\setlength{\tabcolsep}{1.15mm}
\begin{tabular}{llccc}
\toprule
\multirow{3}{*}{\textbf{Method}} & \multirow{3}{*}{\textbf{Ref.}} & \multicolumn{3}{c}{\textbf{AlpacaEval 2}} \\
\cmidrule(lr){3-5}
& & {\small \bf LC (\%)} & {\small \bf WR (\%)} & {\small \bf Len.} \\
\midrule
ORPO      & -       & 10.2 & 7.9  & 1588 \\
Pre-DPO   & ORPO      & 12.3 \small{(+20.6\%)} & 12.1 \small{(+53.2\%)} & 1907 \\
\midrule
sDPO      & last stage       & 12.0 & 11.9  & 1908 \\
Pre-DPO   & sDPO      & \textbf{12.9} \small{(+7.5\%)} & 13.0 \small{(+9.2\%)} & 1951 \\     
\midrule
TR-DPO    & hard update & 11.7 & 12.3 & 1985 \\
Pre-DPO   & TR-DPO    & 12.8 \small{(+9.4\%)} & \textbf{14.2} \small{(+15.4\%)} & 2087  \\
\bottomrule
\end{tabular}
\caption{More results of Pre-DPO with diverse guiding reference models under Llama3.2-3B-Base setting.}
\label{table:other_result_more}
\end{table}

\paragraph{Pre-DPO further improves DPO and SimPO by leveraging guiding reference models.}
In Table~\ref{table:main_result}, we report the performance on the AlpacaEval 2 and Arena-Hard v0.1 across the Llama3.2-3B-Base, Llama3.2-3B-Instruct, Qwen2.5-7B-Base, Qwen2.5-7B-Instruct settings. Compared with baselines, Pre-DPO achieves better performance on the AlpacaEval 2 LC and WR benchmarks, yielding average improvements of 2.5 and 2.8 points, respectively. On the Arena-Hard v0.1 benchmark, Pre-DPO also consistently demonstrates improvements across most settings. For instance, on Qwen2.5-7B-Instruct, Pre-DPO achieves an improved WR of 68.8 compared to 62.9 of the DPO baseline. These results indicate that Pre-DPO is effective in further improving both reference-based and reference-free methods by leveraging guiding reference models.

\paragraph{Pre-DPO improves performance without significantly increasing the response length.}
Although Pre-DPO continuously improves performance, we observe that it does not significantly increase the response length compared to the baselines. Notably, with SimPO as the guiding reference model, Pre-DPO achieves the best performance and the shortest average response length in the Qwen2.5-7B-Base setting.

\paragraph{Pre-DPO is compatible with the iterative preference optimization framework.}
Note that the DPO and SimPO experiments for Llama3.2-3B-Instruct and Qwen2.5-7B-Instruct use on-policy preference datasets constructed by sampling from the current policy, corresponding to the first round of iterative preference optimization~\cite{xiong2024iterative,yuan2024self,rosset2024direct,zhang2025iterative}. Under this setting, Pre-DPO achieves better optimization performance, indicating that Pre-DPO is complementary to the iterative framework and can be employed as part of the iterative process to enhance the use of newly collected preference data.

\subsection{Ablations and More Results}
\paragraph{The guiding reference model plays a critical role in the improvement of Pre-DPO.}

\begin{figure*}[t]
    \centering
    \includegraphics[width=0.98\textwidth]{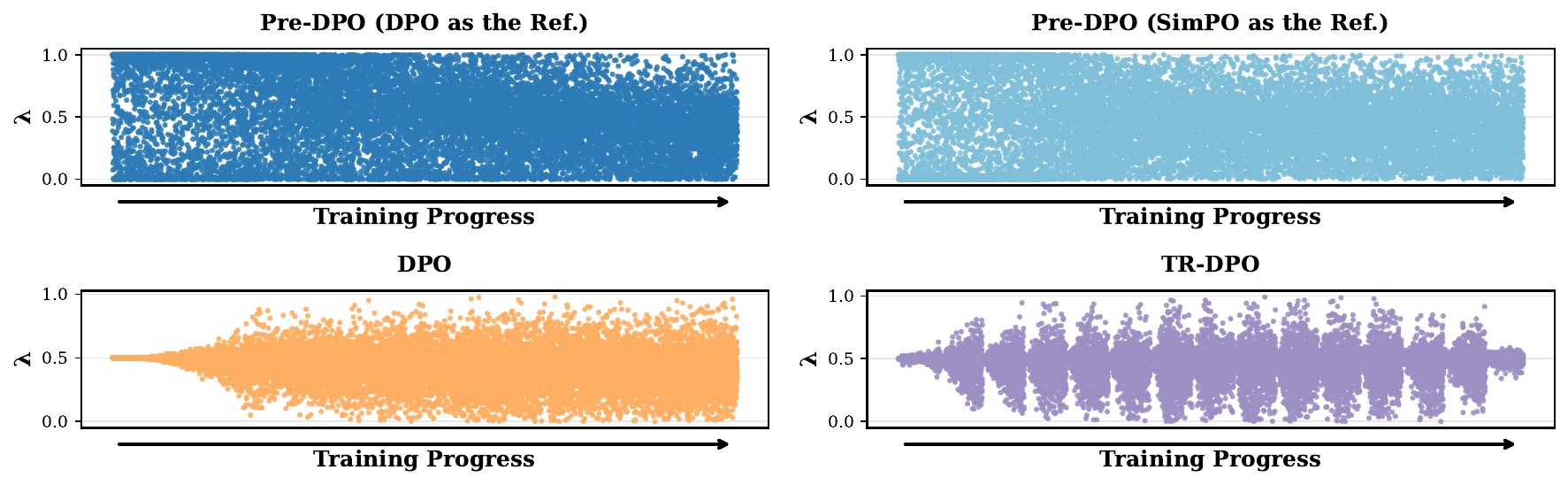}
    \caption{$\lambda$ distribution dynamics of DPO, TR-DPO, and Pre-DPO under the Llama3.2-3B-Base setting. Pre-DPO maintains a broader distribution during the entire training.}
    \label{fig:lambda_comparison}
\end{figure*}

\begin{figure*}[t]
    \centering
    \includegraphics[width=0.98\textwidth]{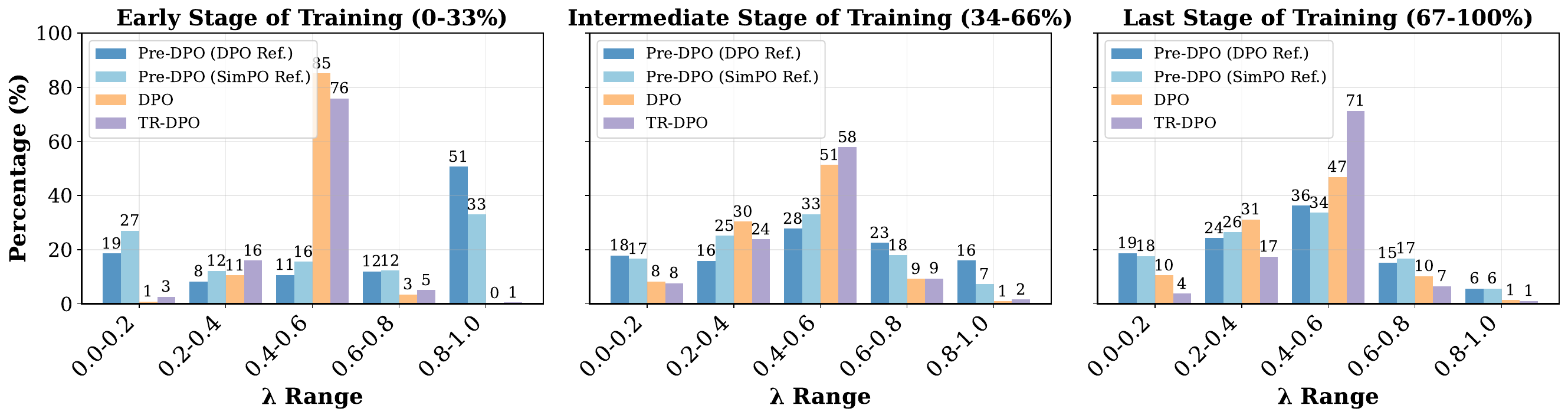}
    \caption{Quantitative analysis of the $\lambda$ distribution during training for DPO, TR-DPO (hard update), and Pre-DPO under the Llama3.2-3B-Base setting. Numerical values on top of the bars indicate the corresponding percentages.}
    \label{fig:lambda_distribution}
\end{figure*}

Although Pre-DPO consistently improves the performance of DPO and SimPO, it introduces additional computational cost due to the need to obtain a guiding reference model. To investigate whether the performance gain is simply due to increased training, we compare Pre-DPO with a baseline where DPO is trained for 2 epochs using the original reference configuration. As shown in Table~\ref{table:ablation_ref}, DPO with a larger computational budget does not yield a noticeable gain, which aligns with previous findings that a single training epoch generally yields the best results~\cite{meng2024simpo}. However, Pre-DPO with a guiding reference model consistently achieves the best LC and WR on AlpacaEval 2, benefiting from the better data utilization enabled by the guiding reference model and avoiding the excessive constraints imposed by traditional reference model setups. 

\paragraph{Pre-DPO can consistently benefit from diverse preference optimization methods.}

To further validate the generality of Pre-DPO, we obtain guiding reference models from more preference optimization methods under the Llama3.2-3B-Base setting, including ORPO, sDPO, and TR-DPO. For sDPO, we divide the preference dataset into two equal parts and perform a two-stage DPO training procedure. For TR-DPO, we adopt a hard update strategy, where the reference model is reset to the current policy every 32 training steps. As shown in Table~\ref{table:other_result_more}, Pre-DPO consistently improves performance on AlpacaEval 2. The results demonstrate that by leveraging guiding reference models, Pre-DPO can consistently benefit from diverse preference optimization methods.

\subsection{The $\lambda$ Distribution during Pre-DPO Training}
Pre-DPO demonstrates a clear advantage via adaptive data reweighting, assigning higher weights to samples aligned with the guiding reference model and down-weighting those that may introduce conflicting learning signals.

To better illustrate the training dynamics, we visualize the evolution of the $\lambda$ distribution throughout training for DPO, TR-DPO, and Pre-DPO under the Llama3.2-3B-Base setting. As intuitively observed in Figure~\ref{fig:lambda_comparison}, Pre-DPO, equipped with a guiding reference model, maintains a broader $\lambda$ distribution throughout training. In contrast, DPO initially assigns nearly uniform data weights and gradually adjusts $\lambda$ to regulate the policy’s deviation from the initial policy. TR-DPO with hard reference updates (i.e., resetting the reference model to the current policy every $k$ steps) tends to maintain a more uniform $\lambda$ distribution throughout training, thereby diminishing the influence of the reference model.

Figure~\ref{fig:lambda_distribution} quantitatively demonstrates that in the early stages of training (0–33\%), Pre-DPO's $\lambda$ values are more concentrated at the extremes (i.e., in the ranges 0–0.2 and 0.8–1.0), whereas those of DPO and TR-DPO are more centered around 0.5. As training progresses into the later stages (34–100\%), Pre-DPO’s $\lambda$ distribution shifts closer to 0.5. Nevertheless, it consistently maintains a broader and more balanced distribution across the entire training.

\section{Conclusion}
We proposed Pre-DPO, a simple yet effective DPO-based preference optimization paradigm that enhances data utilization and improves performance by leveraging a guiding reference model. Unlike traditional DPO, which uses a reference identical to the initial policy, Pre-DPO reuses an optimized policy model as the guiding reference model to re-optimize the initial policy model. This shifts the role of the reference model from a constraint to an informed guide, enabling more effective data reweighting. Extensive experiments across multiple models and scales show that Pre-DPO consistently outperforms both DPO and SimPO, without requiring external models or additional data. We hope this work can inspire more exploration and discussion on the role and improvement of reference models in RLHF for LLMs.

\section*{Acknowledgments}
We would like to sincerely thank all the anonymous reviewers for their valuable feedback. This work is partially funded by the National Natural Science Foundation of China Key Program (Grant No. 62336006).

\bibliography{aaai2026}

\appendix

\section{Derivation of the DPO Gradient}
\label{sec:appendix_derivation}
Given the DPO loss
\begin{align}
    \mathcal{L}_{\text{DPO}}(\pi_{\theta}; \pi_{\text{ref}}) \! = \!
    & -\mathbb{E}_{(x, y^+, y^-) \sim \mathcal{D}} \Big[ \log \sigma \Big( 
    \beta \log \frac{\pi_{\theta}(y^+ \mid x)}{\pi_{\text{ref}}(y^+ \mid x)} \nonumber \\
    & \qquad \qquad - \beta \log \frac{\pi_{\theta}(y^- \mid x)}{\pi_{\text{ref}}(y^- \mid x)} 
    \Big) \Big],
\end{align}
let
\begin{equation}
    z = \beta \log \frac{\pi_\theta(y^+|x)}{\pi_{\text{ref}}(y^+|x)} - \beta \log \frac{\pi_\theta(y^-|x)}{\pi_{\text{ref}}(y^-|x)},
\end{equation}
the gradient can be expressed as
\begin{align}
    \nabla_\theta \mathcal{L}_{\text{DPO}}(\pi_\theta; \pi_{\text{ref}}) 
    &= - \mathbb{E}_{(x, y^+, y^-) \sim \mathcal{D}} \left[ \frac{1}{\sigma(z)} \sigma'(z) \nabla_\theta z \right] \nonumber \\
    &= - \mathbb{E}_{(x, y^+, y^-) \sim \mathcal{D}} \left[ (1 - \sigma(z)) \nabla_\theta z \right] \nonumber \\
    &= - \mathbb{E}_{(x, y^+, y^-) \sim \mathcal{D}} \left[ \sigma(-z) \nabla_\theta z \right] \nonumber \\
    &= - \mathbb{E}_{(x, y^+, y^-) \sim \mathcal{D}} \Big[ \sigma \Big( \beta \log \frac{\pi_{\text{ref}}(y^+|x)}{\pi_{\text{ref}}(y^-|x)} \nonumber \\
    &- \log \frac{\pi_\theta(y^+|x)}{\pi_\theta(y^-|x)} \Big) \nabla_\theta \beta \log \frac{\pi_\theta(y^+ \mid x)}{\pi_\theta(y^- \mid x)} \Big].
\end{align}
We define:
\begin{equation}
    \lambda = \sigma\left( 
        \beta \log \frac{\pi_{\text{ref}}(y^+|x)}{\pi_{\text{ref}}(y^-|x)} 
        - \beta \log \frac{\pi_\theta(y^+|x)}{\pi_\theta(y^-|x)} 
    \right),
\end{equation}
and thus we can obtain
\begin{align}
    \nabla_\theta \mathcal{L}_{\text{DPO}}(\pi_\theta; \pi_{\text{ref}}) = \nonumber \\ 
    & \!\!\!\!\!\!\!\!\!\!\!\!\!\!\!\!\!\!\!\!\!\!\!\!\!\!\!\!\!\! - \beta \mathbb{E}_{(x, y^+, y^-) \sim \mathcal{D}} \left[ 
    \lambda \cdot
    \nabla_\theta \log \frac{\pi_\theta(y^+ \mid x)}{\pi_\theta(y^- \mid x)}
    \right].
\end{align}

\section{More Experimental Details}
\subsection{Training Hyperparameters}
\label{sec:appendix_implementation_details}

\paragraph{Hyperparameter tuning.}
Considering that hyperparameter tuning is crucial for preference optimization, we adopt a two-stage hyperparameter search method to ensure both fairness and efficiency. Specifically, in the first stage, we fix the learning rate at $6 \times 10^{-7}$ and individually search for the optimal $\beta$ in $[0.005, 0.01, 0.05, 0.1, 0.2, 0.5, 1.0]$ for DPO-based methods, search for $\beta$ in $[2.0, 2.5]$ and $\gamma$ in $[0.3, 0.5, 1.0, 1.2, 1.4, 1.6]$ for SimPO, and search for $\lambda$ in $[0.1, 0.5, 1.0, 2.0, 3.0, 5.0]$ for ORPO. In the second stage, we select the two best hyperparameter settings found in the first stage and individually search for the optimal learning rate in $[3 \times 10^{-7}, 5 \times 10^{-7}, 8 \times 10^{-7}, 1 \times 10^{-6}]$. The search ranges for the hyperparameters are chosen with reference to prior work~\cite{meng2024simpo}.

\paragraph{The optimal hyperparameter values.}

\begin{table*}[t]
\centering
\begin{tabular}{lccccc}
\toprule
\textbf{Policy Model} & \textbf{Method} & \textbf{Ref.} & $\boldsymbol\beta$ & $\boldsymbol\gamma$ & \textbf{Learning Rate} \\
\midrule
Llama3.2-3B-Base-SFT  & DPO       & SFT   & 0.005 & -   & $1 \times 10^{-6}$ \\
Llama3.2-3B-Base-SFT  & SimPO     & SFT   & 2.5   & 1.2 & $1 \times 10^{-6}$ \\
Llama3.2-3B-Base-SFT  & Pre-DPO & DPO   & 0.05  & -   & $1 \times 10^{-6}$ \\
Llama3.2-3B-Base-SFT  & Pre-DPO & SimPO & 0.05  & -   & $6 \times 10^{-7}$ \\
Llama3.2-3B-Instruct  & DPO       & SFT   & 0.05  & -   & $6 \times 10^{-7}$ \\
Llama3.2-3B-Instruct  & SimPO     & SFT   & 2.5   & 1.0 & $1 \times 10^{-6}$ \\
Llama3.2-3B-Instruct  & Pre-DPO & DPO   & 0.05  & -   & $1 \times 10^{-6}$ \\
Llama3.2-3B-Instruct  & Pre-DPO & SimPO & 0.1   & -   & $1 \times 10^{-6}$ \\
Qwen2.5-7B-Base-SFT   & DPO       & SFT   & 0.005 & -   & $8 \times 10^{-7}$ \\
Qwen2.5-7B-Base-SFT   & SimPO     & SFT   & 2.5   & 1.4 & $8 \times 10^{-7}$ \\
Qwen2.5-7B-Base-SFT   & Pre-DPO & DPO   & 0.2   & -   & $8 \times 10^{-7}$ \\
Qwen2.5-7B-Base-SFT   & Pre-DPO & SimPO & 0.2   & -   & $1 \times 10^{-6}$ \\
Qwen2.5-7B-Instruct   & DPO       & SFT   & 0.01  & -   & $5 \times 10^{-7}$ \\
Qwen2.5-7B-Instruct   & SimPO     & SFT   & 2.5   & 1.2 & $1 \times 10^{-6}$ \\
Qwen2.5-7B-Instruct   & Pre-DPO & DPO   & 0.05  & -   & $1 \times 10^{-6}$ \\
Qwen2.5-7B-Instruct   & Pre-DPO & SimPO & 0.1   & -   & $6 \times 10^{-7}$ \\
\bottomrule
\end{tabular}
\caption{Optimal hyperparameters for the main experiments.}
\label{table:hyperparams}
\end{table*}

Table~\ref{table:hyperparams} shows the optimal hyperparameter values in our main experiments. It can be observed that Pre-DPO generally requires a larger $\beta$ compared to DPO.

\subsection{Evaluation Details}
\label{sec:appendix_evaluation_details}
Following previous work~\cite{meng2024simpo}, we adopt a sampling-based decoding strategy for AlpacaEval 2 with a temperature of 0.7, a top-$p$ of 0.9, and a greedy decoding strategy for Arena-Hard v0.1. We evaluate the AlpacaEval 2 and Arena-Hard v0.1 results using their official repositories, both of which employ gpt-4-1106-preview as the judge model.

\section{More Analysis}
\subsection{Response Length Differences of $y^{+/-}$ Across Datasets}
It can be observed that SimPO shows a significant advantage over DPO in the Base settings, while DPO slightly outperforms SimPO in the Instruct settings. A possible reason is that the preference datasets constructed via on-policy sampling exhibit smaller length differences between positive responses $y^+$ and negative responses $y^-$, thus diminishing the advantage of SimPO's length normalization.

We define the normalized length difference between positive and negative responses in preference datasets as follows:
\begin{equation}
\text{Normalized Length Difference} = \frac{|\text{len}(y^+)-\text{len}(y^-)|}{\max \left(\text{len}(y^+), \text{len}(y^-) \right)},
\end{equation}
where $\text{len}(y)$ denotes the number of tokens in response $y$.

We compute this metric on three datasets: the original UltraFeedback preference data used for Base models, and the constructed UltraFeedback-Llama3.2-3B-Instruct and UltraFeedback-Qwen2.5-7B-Instruct preference datasets used for Instruct models. The original UltraFeedback dataset exhibits an average normalized length difference of 0.465, while the resampled on-policy datasets, UltraFeedback-Llama3.2-3B-Instruct and UltraFeedback-Qwen2.5-7B-Instruct, have smaller values of 0.296 and 0.224, respectively. This indicates that the resampled datasets contain less variation in response length, potentially imposing more stringent optimization conditions for SimPO.

\subsection{One Iteration of Pre-DPO Can Be Enough}

\begin{table*}[t]
\centering
\begin{tabular}{lccccc}
\toprule
\multirow{3}{*}{\textbf{Method}} & \multirow{3}{*}{\textbf{Initial Policy}} & \multirow{3}{*}{\textbf{Ref.}} & \multicolumn{3}{c}{\textbf{AlpacaEval 2}} \\
\cmidrule(lr){4-6}
& & & {\small \bf LC (\%)} & {\small \bf WR (\%)} & {\small \bf Len.} \\
\midrule
DPO      & SFT & SFT & 10.5 & 12.0 & 2042 \\
Pre-DPO \scriptsize{round 1}  & SFT & DPO & \textbf{12.5} & 13.9 & 2061 \\
Pre-DPO \scriptsize{round 2}  & SFT & Pre-DPO \scriptsize{round 1} & 12.2 & \textbf{14.0} & 2134  \\
\bottomrule
\end{tabular}
\caption{A further round of Pre-DPO under the Llama3.2-3B-Base setting. With sufficient hyperparameter tuning, one round of Pre-DPO can be enough.}
\label{table:appendix_one_iteration_enough}
\end{table*}

The purpose of the guiding reference model—whether derived from DPO or SimPO—is not to directly achieve optimal performance, but to facilitate more effective use of existing preference data and guide subsequent preference optimization. Additional empirical results indicate that, with proper hyperparameter tuning, a single iteration of Pre-DPO is generally sufficient for effective data utilization. As shown in Table~\ref{table:appendix_one_iteration_enough}, applying Pre-DPO for an additional iteration yields negligible improvements on AlpacaEval 2 in the Llama3.2-3B-Base setting.

\end{document}